\documentclass[default,iicol]{sn-jnl}

\jyear{2021}%

\theoremstyle{thmstyleone}%
\theoremstyle{thmstyletwo}%

\theoremstyle{thmstylethree}%

\usepackage{mathtools}
\begin{document}

\title[Article Title]{TopTrack: Tracking Objects By Their Top}


\author*[1]{\fnm{Jacob} \sur{Meilleur}}\email{jacob.meilleur@polymtl.ca}

\author[1]{\fnm{Guillaume-Alexandre} \sur{Bilodeau}}\email{gabilodeau@polymtl.ca}

\affil*[1]{\orgdiv{LITIV}, \orgname{Polytechnique Montréal}, \orgaddress{\street{2500, chemin de Polytechnique}, \city{Montréal}, \postcode{H3T 1J4}, \state{Québec}, \country{Canada}}}


\abstract{In recent years, the joint detection-and-tracking paradigm has been a very popular way of tackling the multi-object tracking (MOT) task. Many of the methods following this paradigm use the object center keypoint for detection. However, we argue that the center point is not optimal since it is often not visible in crowded scenarios, which results in many missed detections when the objects are partially occluded. We propose \textit{TopTrack}, a joint detection-and-tracking method that uses the top of the object as a keypoint for detection instead of the center because it is more often visible. Furthermore, \textit{TopTrack} processes consecutive frames in separate streams in order to facilitate training. We performed experiments to show that using the object top as a keypoint for detection can reduce the amount of missed detections, which in turn leads to more complete trajectories and less lost trajectories. \textit{TopTrack} manages to achieve competitive results with other state-of-the-art trackers on two MOT benchmarks.}




\maketitle

\section{Introduction}\label{sec1}

Multi-object Tracking (MOT) is a hallmark task of computer vision that consists of detecting all the objects of predefined classes in each frame of a video and linking together the detections of the same object through time in order to create a trajectory for each object. Most state-of-the-art (SOTA) online trackers follow the tracking-by-detection paradigm, where the tracking task is divided into two subtasks: object detection and data association. Firstly, objects are detected in each frame and then the objects are associated with each other in a separate step using mostly motion and appearance cues extracted from each frame or group of frames. This method works well, but usually requires a high computational cost and often relies on complex data association methods. Because of this, recent works have introduced the joint-tracking-and-detection paradigm where the tracking task is reconciled into a single framework. These models manage to achieve competitive results while being both simpler and faster than models following the tracking-by-detection paradigm. \textit{CenterTrack} \cite{TrackingObjectsAsPoints} is such a model. It represents objects by their center point to perform detection and tracking and then regresses the bounding boxes from these center points in order to extract the position and size of every object. \textit{CenterTrack} demonstrated that tracking objects at the point-level simplifies two core components of MOT. Firstly, it allows for the representation of the candidate objects as a heatmap of points, which can then be fed as input to the network so that it can reason about every object in the frame jointly with those of the previous frame. Secondly, it simplifies data association, which can be performed with a simple displacement prediction that is conditioned on prior detections. 

However, \textit{CenterTrack} also has some drawbacks. Indeed, occluded objects remains one of MOT biggest problem to date, even with the introduction of the joint-tracking-and-detection paradigm. This is especially true for crowded scenes, where there are many partially or fully occluded objects. In these scenes, the center of the objects is often not visible, which can affect the performance of \textit{CenterTrack} and other similar trackers, since they rely on the center point of the objects for detection. This is sometimes alleviated using the heatmap of the previous frame as input to the network to allow it to carryover some detections from one frame to the other, even if the object is not visible in the current frame. However, this is often not the case and detections are missed because the center of the objects is not visible while the rest is.

In this paper, we propose a model that uses the top of the objects as a keypoint for detection instead of the center because it is more often visible in these scenarios, which in turn allows for better detection results. Moreover, we propose a simpler and more intuitive architecture that does not rely on the generated heatmap as input for the whole network while training, but instead only uses it for tracking purposes. We also show how using a combination of the current frame, the previous frame, the current heatmap and the previous heatmap for tracking allows our method to better connect objects locally.

We evaluated our method on the MOT17 challenge benchmark \cite{MOT17} and the MOT20 challenge benchmark \cite{MOT20} and we show how our proposed method manages to better detect hard-to-track objects, resulting in more complete trajectories and less lost tracks. Our contributions are summarized as follows:
\begin{itemize}
    \item We introduce \textit{TopTrack}, an end-to-end trainable multi-object tracker that uses the top of the objects for detection and tracking. We show that using this keypoint instead of the center of the objects is beneficial for this task.
    \item We show that our method can achieve competitive tracking accuracy on popular MOT datasets: MOT17 and MOT20.
\end{itemize}

\section{Related Work}

Most SOTA trackers fall under the tracking-by-detection paradigm \cite{ByteTrack, C-BIOU, bot-sort, StrongSORT}, which is characterized by performing the object detection and data association in two distinct stages, and is often accomplished using two distinct models. An off-the-shelf object detector \cite{FasterR-CNN, DPM, SDP, YOLOX, CascadeRCNN} is used in order to get candidate bounding boxes for each frame. Then, data association can be performed between the bounding boxes of each subsequent frame in order to link the identities of each object through time. The data association stage often makes use of motion and appearance cues in order to achieve better results. Commonly used motion cues are motion prediction models like Kalman filters \cite{KalmanFilters, NSAKalman} and camera motion compensation \cite{ECC}. Appearance cues include ReID features obtained from a pre-trained model \cite{BoT, CNN_reid, LuNet} or learned jointly \cite{SiameseCNN, JDE, FairMOT}. ReID features can also be augmented with other visual information like pose estimation \cite{LMP}.

Examples of tracking-by-detection methods include \textit{SORT} \cite{SORT} that uses \textit{Faster R-CNN} \cite{FasterR-CNN} to perform object detection and uses Kalman filters \cite{KalmanFilters} to predict the motion of each object. Data association is then performed between the prediction of the Kalman filters and the detector using Intersection over Union (IoU) as a distance metric and the Hungarian algorithm for matching \cite{HungarianMatching}. \textit{DeepSORT} \cite{DeepSORT} adds appearance information and cascade matching in the data association stage of \textit{SORT} in order to improve results. \textit{StrongSORT} \cite{StrongSORT} updates the components of \textit{DeepSORT} to better performing ones and introduces a new global tracklet linking module and a Gaussian smoothing interpolation module to achieve SOTA results. \textit{SST}~\cite{SST} proposes a Deep Affinity Network, which models the appearance of the objects and does the data association by computing the affinity between the existing tracks and the detected object and then matching them using the Hungarian algorithm~\cite{HungarianMatching}.  \textit{C-BIOU} \cite{C-BIOU} uses detections from \textit{YOLOX} \cite{YOLOX} then performs cascade matching between existing tracklets and the detections. The size of the bounding boxes is enlarged by a constant to allow for the tracking of hard examples where the position of the object has changed a lot. The first matching uses a small enlargement factor, then the remaining unmatched tracks and detections are matched again using a large enlargement factor. \textit{ByteTrack} \cite{ByteTrack} proposes a data association strategy that focuses on matching all detections extracted from an off-the-shelf object detector, even those with a low confidence score instead of ignoring them. This is done by first matching the detection with a high confidence score, determined by a fine-tuned threshold, with the existing tracks. Then, the detections with lower confidence scores are matched with the remaining tracks. 

Many recent trackers \cite{TrackingObjectsAsPoints,FairMOT,Tracktor,TrackFormer,TransCenter,Transtrack, MOTR, ChainedTracker,GSDT, TubeTK} are using a novel paradigm, where the detection and tracking are performed jointly. These methods are characterized by their use of tracking information in order to boost the detection results in future frames and by using a single model in order to make the predictions for both detection and tracking. Then, a post-processing step will use these predictions in order to obtain the final results. \textit{TubeTK}~\cite{TubeTK} and \textit{Chained-Tracker}~\cite{ChainedTracker} pioneered the concept of joint-detection-and-tracking by proposing a completely end-to-end trainable pipeline. \textit{TubeTK} proposes the concept of \emph{bounding-tubes} to represent the spatiotemporal location of objects detected in a video. A \emph{bounding-tube} is defined as a series of three bounding boxes of the same object from different frames. Tracks are broken down into a combination of tubes where the bounding box at each frame is defined as the middle bounding box of a tube in order to link tubes together. The three bounding boxes forming a tube do not have to be from consecutive frames, which allows the interpolation of the object location within a tube. Tubes are linked using IoU as a distance metric. \textit{Chained-Tracker} takes two adjacent frames as input and generates detection pairs of each object in both frames and then link each consecutive pair using IoU as a distance metric and then the Hungarian algorithm~\cite{HungarianMatching} for matching.

\textit{CenterTrack}~\cite{TrackingObjectsAsPoints} learns to generate a center heatmap for object detection and an offset vector that represents the displacement from one frame to the next using the two consecutive frames and the center heatmap of the previous frame. It then regresses the bounding boxes from the center points of the objects. Data association is done using greedy matching between the position of the objects in the previous frame and the position of the predicted offset. \textit{Tracktor}~\cite{Tracktor} exploits the bounding box regression module of \textit{Faster R-CNN} to perform tracking directly from the information provided by the detector. Moreover, the model can be extended with a motion model and a re-identification algorithm to achieve better results. \textit{FairMOT}~\cite{FairMOT} aims to reconcile the inherent bias in favor of the detection task when training a MOT model. Indeed, they noted three ways in which trackers following the joint detection-and-tracking paradigm are biased towards the detection task, which harms the global results. As a result, they use \textit{CenterNet}~\cite{ObjectsAsPoints} for their detection branch and add a parallel ReID feature embedding branch to generate robust appearance features that are then used during the data association step to improve their tracking results. \textit{GSDT}~\cite{GSDT} proposes to use graph neural networks to extract object features in order to model object-object relation within the frame. This is done by creating a graph where the nodes are the features of the detected objects and of the tracklets and the edges are formed between every detection node within a pre-defined spatial window of a tracklet node. Node features are first obtained using \textit{CenterNet} and are then iteratively updated using the features of its neighbors. Finally, they are used for data association in conjunction with learned identity embeddings. 

Several models using transformers for feature extraction~\cite{TransCenter,Transtrack,MOTR} have also been proposed. These methods have shown to be accurate, but demand a very high amount of computational resources. \textit{TransCenter} \cite{TransCenter} follows \cite{ObjectsAsPoints,TrackingObjectsAsPoints,GSDT} by detecting object from their center point using a learned generated heatmap and introduces a query learning network (QLN) that learns dense queries in order to allow the transformer architecture to produce dense representations. \textit{MOTR}~\cite{MOTR} extends \textit{DETR}~\cite{DETR} to the MOT task by changing \textit{DETR} fixed-length object queries set to a track queries set that is dynamically updated and that has a variable length. Furthermore, a Tracklet-Aware Label Assignment strategy is introduced to better perform data association. \textit{DETR} bipartite matching strategy is kept for new tracks, but existing tracks are matched according to their track query.

\begin{figure*}[h]
\centering
\includegraphics[width=0.8\textwidth]{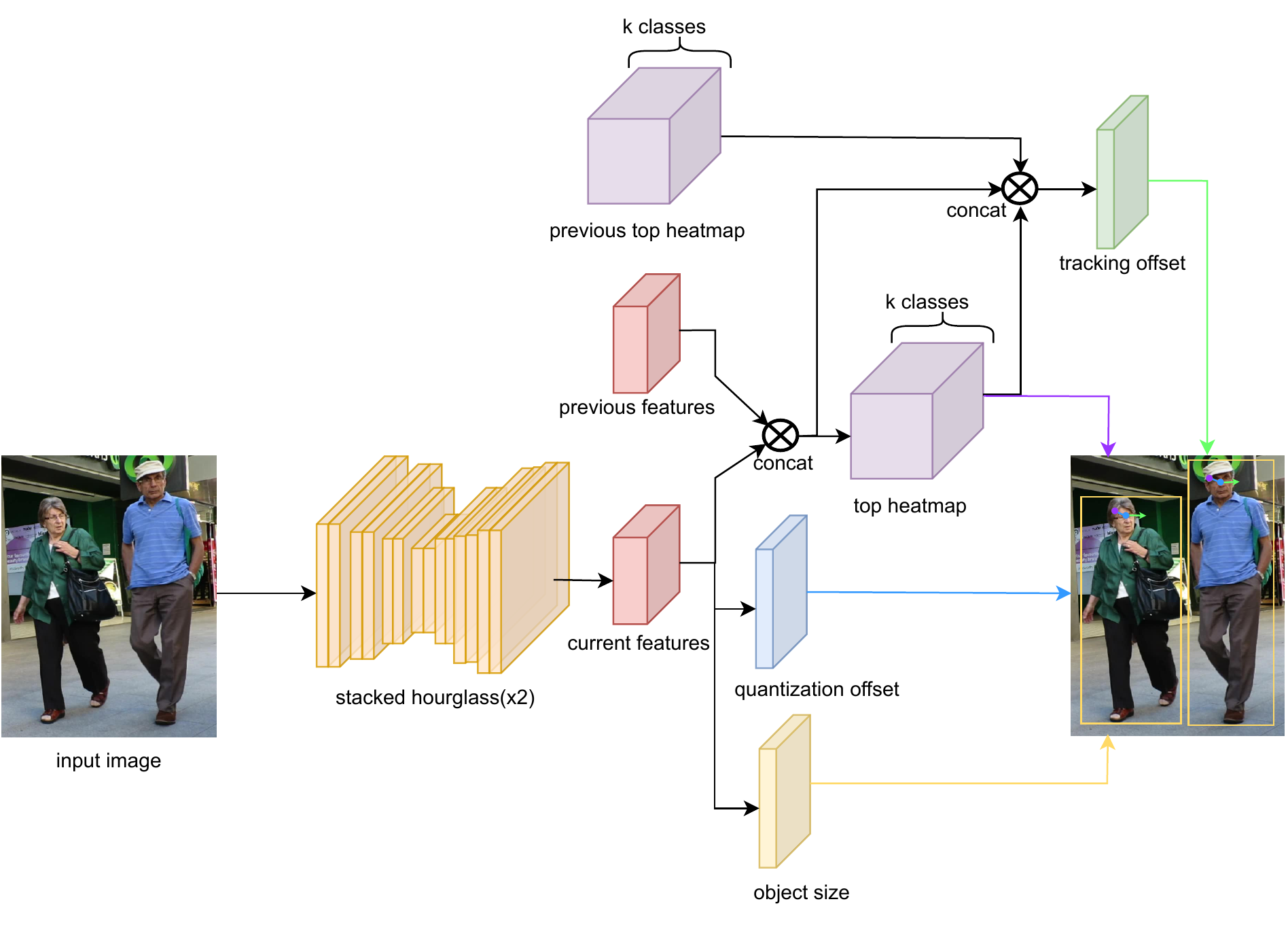}
\caption{Overview of the architecture of TopTrack: An image at timestep $t$ is fed to a Stacked Hourglass backbone to extract the feature map. This feature map is used to predict the size of the objects and the quantization offset. The feature map from the previous timestep $t-1$ is saved and then used in conjunction with the current feature map to generate the top heatmaps. Finally, the current feature map, the current top heatmaps, the previous feature map and the previous top heatmaps are used to predict the tracking offset}
\label{fig:architecture}
\end{figure*}

\section{Proposed Method}\label{sec11}
We propose a method based on CNNs, making it more efficient and tractable than with Transformers. It uses a simpler and more intuitive architecture than \textit{CenterTrack} \cite{TrackingObjectsAsPoints} by not relying on the previous generated heatmap $\hat{H}^{t-1}$ as input for the whole network to facilitate training. Furthermore, we hypothesize that using the top instead of the center point for detection increases accuracy in crowded scenes since the head is more often visible than the center of the body which leads to fewer missed detections. Our proposed model uses a CNN backbone and adds other prediction heads on top of it in order to perform object detection and multi-object tracking. We call our proposed method, \textit{TopTrack}.

As shown in Figure \ref{fig:architecture}, \textit{TopTrack} only takes the current image $I^t \in \mathbb{R}^{H \times W \times 3}$ as input. The saved previous feature map $F^{t-1} \in \mathbb{R}^{H \times W \times 3}$ and previous heatmap $\hat{H}^{t-1} \in \mathbb{R}^{\frac{H}{R} \times \frac{W}{R} \times C}$ are also used by some network heads. The image $I^t$ is first sent to a CNN backbone in order to extract the feature map $F^t \in \mathbb{R}^{\frac{H}{R} \times \frac{W}{R} \times 3}$, with $R=4$ being a downsampling factor. 

\begin{figure}
    \centering
    \includegraphics[width=0.5\textwidth]{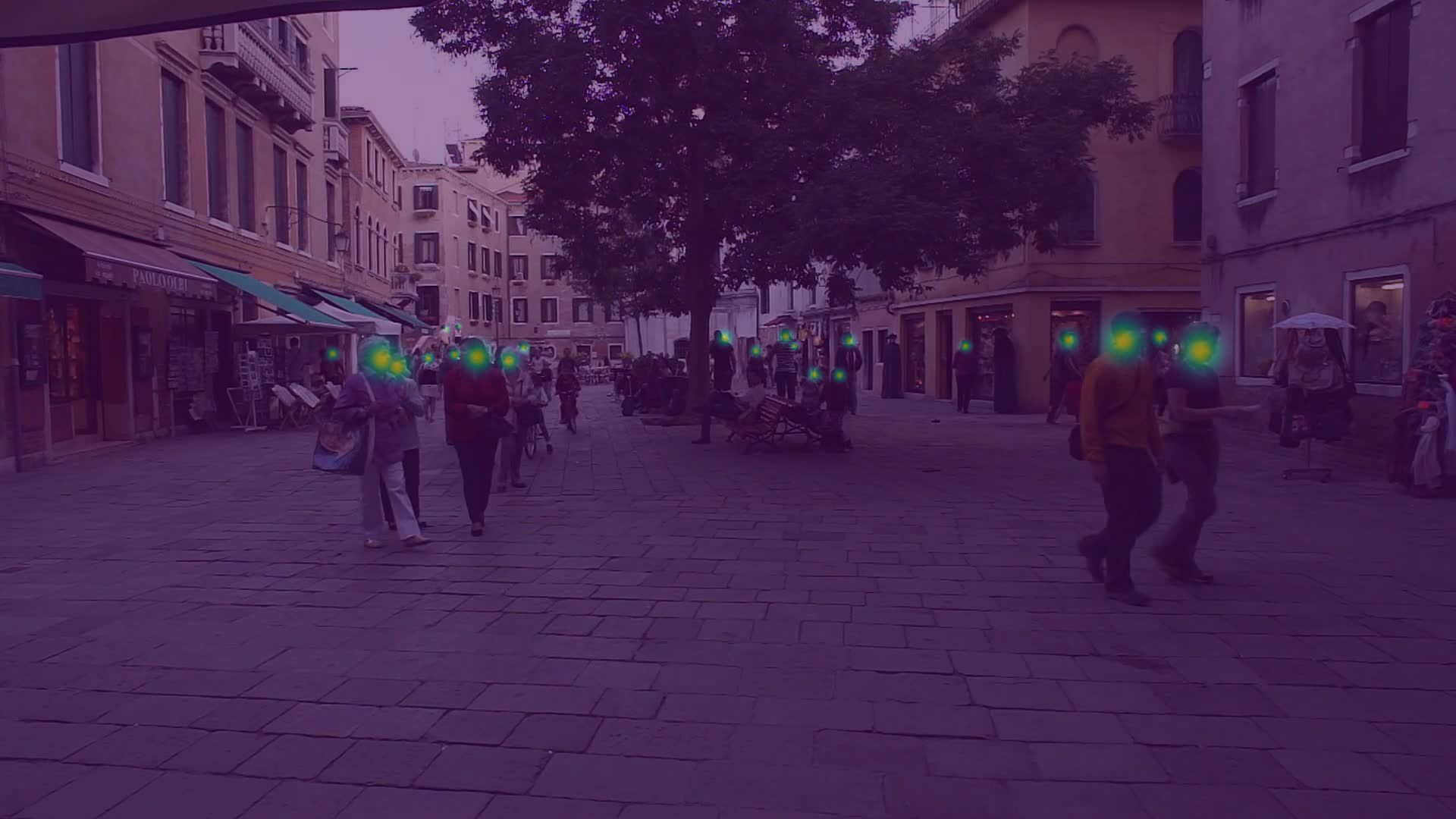}
    \caption{Visualization of how top heatmaps are used to extract object position. The likelihood of a certain pixel being the position of the top of an object is represented by a Gaussian function. The positions are extracted by identifying the peaks of each Gaussian function}
    \label{fig:heatmap}
\end{figure}

Then, the feature map $F^t$ is used in order to predict the size of the bounding box $\hat{S}^t \in \mathbb{R}^{\frac{H}{R} \times \frac{W}{R} \times 2}$ for each object and a quantization offset $\hat{O}^t \in \mathbb{R}^{\frac{H}{R} \times \frac{W}{R} \times 2}$. $F^t$ and $F^{t-1}$ are used to generate a top point heatmap $\hat{H}^t \in \mathbb{R}^{\frac{H}{R} \times \frac{W}{R} \times C}$ for each pre-defined object class $C$. Finally, $F^t$, $F^{t-1}$, $\hat{H}^t$ and $\hat{H}^{t-1}$ are used to predict the displacement $\hat{D}^t \in \mathbb{R}^{\frac{H}{R} \times \frac{W}{R} \times 2}$ between frames  of each object. Each network head of \textit{TopTrack} makes a prediction at every pixel of the feature map. The position of the objects is extracted by identifying the peaks in the generated top point heatmap (see Figure \ref{fig:heatmap}). Then, the correct predictions from the other heads are extracted using the predicted position of the top point of each object. 

\subsection{Top Heatmap}
A first network head generates a top point heatmap $\hat{H}^t$ using the feature maps $F^t$ and $F^{t-1}$ where each element $\hat{h} = [0,1] \in \mathbb{R} $ represents the probability of a given pixel being the top point of an object at this position in $F^t$. Using $F^{t-1}$ helps the detection capabilities of the network by providing additional context about the objects. Essentially, the heatmap is a collection of Gaussian distributions where the highest values (peaks) represent the position of the top of the objects. Extracting the location of every object simply requires taking the peak of each Gaussian distribution. In practice, an element is considered a peak if its value is greater or equal to its eight neighbors in order to account for overlapping objects. During inference, the top-100 peaks are extracted. 

Since only the bounding box positions are given as annotations, we have to generate our own top position ground truths (GT) for training. According to our observations, the top point of an object is usually positioned at half the width and $\frac{1}{10}$ of the height of the bounding box. This is particularly true for humans, but this also captures the top area of other objects. Therefore, given a GT bounding box b = $(x_1, y_1, w, h)$, the ground truth top position is $p = (x, y)$ where $x = x_1 + \frac{w}{2}$ and $y = y_1 + \frac{h}{10}$. Taking the downsampling factor into account, the position of the top becomes $\Tilde{p}=[\frac{p}{R}]$. The GT heatmap $H$ is then created by turning the GT top positions into Gaussian distributions using a Gaussian kernel. The value of the pixel $H_{xy}$ at coordinates $(x,y)$ in $H$ is computed as $H_{xy} = \exp{(-\frac{(x-\Tilde{p}_x)^2 + (y-\Tilde{p}_y)^2}{2\sigma_p^2})}$, where $\sigma_p$ is the standard deviation \cite{CornerNet}. A penalty-reduced pixel-wise logistic regression with focal loss \cite{FocalLoss} is used as the loss function of this network head \cite{ObjectsAsPoints}. Given a positive example ($H_{xy}=1$), the corresponding loss is defined as 
\begin{equation}
    l = (1-\hat{H}_{xy})^{\alpha}\log{(\hat{H}_{xy})}.
\end{equation}

Otherwise, given a negative example, the corresponding loss is defined as
\begin{equation} \label{negative_focal_loss}
    l=(1-H_{xy})^{\beta}(\hat{H}_{xy})^{\alpha} \log{(1-\hat{H}_{xy})}.
\end{equation}

$\alpha$ and $\beta$ are hyper-parameters of the focal loss, set to 2 and 4 respectively following \cite{ObjectsAsPoints,CornerNet}. In \eqref{negative_focal_loss}, a penalty-reduction coefficient is added to the original focal loss equation so that positions near the GT are less penalized. The overall loss then becomes the sum of all examples normalized by the number of objects $N$ and is given by

\begin{equation}
    L_{h} = \frac{-1}{N} \sum\limits_{xy} l.
\end{equation}

\subsection{Object Size}
This network head uses $\hat{I}^t$ to predict the height and width of the bounding box $\hat{S}^t_i$ of every detected object $i$. It makes a prediction $\hat{b} = (\hat{b}_w, \hat{b}_h)$ at every pixel of $\hat{I}^t$, but only the pixels corresponding to the top of an object will be extracted using the generated heatmap $\hat{H}^t$. Given a predicted bounding box size $(\hat{b}_w,\hat{b}_h)$ and a top point $(p_x,p_y)$, the final bounding box coordinates can be computed as $b = (p_x - \frac{\hat{b}_w}{2}, p_y - \frac{\hat{b}_h}{10}, p_x + \frac{\hat{b}_w}{2}, p_x + \frac{9\hat{b}_h}{10})$. For training, the GT bounding boxes are provided under the format $b = (x_1,y_1,x_2,y_2)$. The GT size is then computed as $s = (x_2-x_1,y_2-y_1)$. The training is done using an L1 loss function conditioned on the size predictions extracted from the top point locations only. Let $\hat{S}_i$ be the predicted bounding box at the top position of object $i$, and $s_i$ be the GT bounding box of object $i$, the loss is

\begin{equation}
    L_{size} = \frac{1}{N}\sum\limits_{k=1}^N \lvert \hat{S}_{i} - s_i \rvert .
\end{equation}

\subsection{Quantization offset}

Downsampling the input image by a factor of $R$ has the side effect of introducing a quantization error due to the discretisation of the top point position. As a result, it is possible for the top position of an object to be off by up to $R^2$ pixels, which can have a significant effect on both the detection and the data association tasks. In order to alleviate the quantization error, this network head predicts an offset $\hat{O}$ at each pixel of $\hat{I}^t$. Given the GT top point $p = (p_x,p_y)$ of an object downsampled by a factor of $R$ and its discrete position $\Tilde{p}=(\lfloor p_x \rfloor, \lfloor p_y \rfloor)$, the GT offset of object $i$ is computed as $O_i = p - \Tilde{p}$. The offset loss supervised at the detected top positions of each object $i$ is

\begin{equation}
    L_{off} = \frac{1}{N} \sum\limits_{k=1}^N \lvert \hat{O}_{i} - O_i \rvert .
\end{equation}

\subsection{Displacement Offset}

The model learns to predict the displacement $\hat{D}^t$ of each object $i$ between the current frame $I^t$ and the previous frame $I^{t-1}$. A prediction is made at each pixel of $\hat{I}^t$ and the correct prediction for each object is extracted using their top point position detected by the top heatmap $\hat{H}^t$. This predicted displacement can then be used for data association. 

Given the GT position of object $i$ in the current frame $p^t_i$ and its GT position in the previous frame $p^{t-1}_i$, the GT displacement is computed as $D^t_i = p^t_i - p^{t-1}_i$. The displacement loss is
\begin{equation}
    L_D = \frac{1}{N} \sum\limits_{k-1}^N \lvert \hat{D}_i - D_i \rvert .
\end{equation}

\subsection{Data Association}
We used the same data association step as \textit{CenterTrack}. It consists of a simple greedy association algorithm based on the distance $d = \hat{P}^{t-1} - \tilde{P}^{t-1}$ between the detected position of the top point $\hat{P}^{t-1} = \{p_i\}_{i=0}^{N-1}$ of each of the $N$ objects in the previous frame $I^{t-1}$ and the predicted position of the top point $\tilde{P}^{t-1} = \hat{P}^t - \hat{D}^t$ of each object based on their detected position $\hat{P}^t$ in the current frame and their predicted displacement $\hat{D}^t$. Detections in $I^t$ are matched with the closest existing track. If a match was correctly made, the position of the track is updated with the position of the detection. Detections that remain unmatched are used to create new tracks and tracks that were not matched are deleted. This algorithm is local only, meaning that it does not allow for track rebirths or global association. Using the same association algorithm as \textit{CenterTrack} ensures that the differences in our results are solely due to the changes in the architecture of the network and the usage of the top instead of the center point for object detection.

\section{Training}
\label{Training}
\textit{TopTrack} is end-to-end trainable by combining the losses of each network head into a global training objective. A scaling constant $\lambda$ is used to adjust the weight on the overall training objective if needed. The training objective of the network becomes 
\begin{equation}
    L =  L_h + \lambda_{size} L_{size} + L_{off} +  L_D,
\end{equation}
where the scaling constant $\lambda_{size}$ is set to $0.1$ for all experiments.

In order for the model to be trained properly, strong data augmentation techniques have to be used. Else, the heatmaps generated by the model, which are essential for successful tracking, are subpar. Also, if a video has a high frame rate, the object displacement between frames is very small, so the network could learn to repeat its previous predictions~\cite{TrackingObjectsAsPoints}. These issues are solved by using different data augmentation techniques and by adding common errors made at test time, which includes adding random noise in the GT heatmaps using a Gaussian distribution, removing GT detections to create false negatives and adding false detections to create false positives, flipping the input image horizontally, cropping the input image and modifying its color. All these data augmentation techniques are applied randomly during the training process. Additionally, the previous heatmap $H^{t-1}$ is provided to the model and it is modified using the same techniques as the input image $I^t$ for consistency. Furthermore, during training, the exact previous image $I^{t-1}$ is not always used. Indeed, it is possible to use another image temporally close to the current frame that is not the immediate previous image. $I^{t-1}$ is chosen randomly from an interval of frames $[I^{t-k}, I^{t+k}]$, where $0 < k \leq 3$. This helps to alleviate overfitting to video framerate and unbalanced movement direction in the dataset. To train our model properly using this technique, the features of the previous image $\hat{I}^{t-1}$ have to be extracted alongside those of the current image instead of being able to save them at each timestep as in inference, because the same data augmentation has to be applied to both.

Following \textit{CenterTrack}~\cite{TrackingObjectsAsPoints}, \textit{TopTrack} can be trained on static images. Although the previous frame $I^{t-1}$ does not exist in this case, one can be simulated and used for training by randomly scaling and translating the current frame $I^t$. 

\section{Experiments}

We used three datasets to train \textit{TopTrack}: MOT17~\cite{MOT17}, MOT20~\cite{MOT20} and CrowdHuman~\cite{CrowdHuman} and evaluated it on MOT17 and MOT20. These datasets do not provide official validation sets so the training sets were split in half in order to allow hyperparameter tuning and to perform the ablation study of section~\ref{Ablation Study}. The final results were obtained by training \textit{TopTrack} on the whole training set of either MOT17 or MOT20 and submitting the results to the respective test server of each dataset for evaluation. 

The MOT17 challenge dataset contains 14 video sequences, 7 for training and 7 for testing. Many considerations were taken when crafting this dataset in order to better evaluate MOT models. Firstly, in some sequences the camera is moving and in others it is static. It can also be positioned at a high, medium or low height to have different viewpoints. Moreover, sequences are filmed in different weather to account for different illumination conditions (sunny, cloudy, night). Finally, the crowd density and number of occlusions are much higher than in MOT15~\cite{MOT15}. MOT17 is an extension of MOT16 that provides public detections with more detectors and more precise annotations. 

The MOT20 challenge is a more recent dataset created to tackle scenes that have denser crowds compared to its predecessor. It contains 4 training sequences and 4 test sequences, taken from 3 very crowded scenes that can go up to a mean density of 246 pedestrians per frame. The test set contains two sequences taken from the same scenes as the sequences from the training set and two sequences that are completely new to the model. This dataset includes indoor and outdoor sequences, as well as daytime and nighttime settings. All sequences are filmed from an elevated point-of-view.

The CrowdHuman dataset is an object detection dataset that is used to pre-train \textit{TopTrack}. It is a human detection benchmark that is characterized by its crowded scenes and high number of occlusions. Using strong data augmentation allows for the training of a MOT model even though it is a single image object detection dataset as described at the end of section~\ref{Training}. 

Our model is evaluated using all the standard MOT metrics, such as the CLEAR MOT metrics (MOTA, MOTP, MT, etc)~\cite{MOTA}, IDF1~\cite{IDF1} and HOTA~\cite{HOTA}.

\subsection{Implementation Details} \label{Implementation Details}
\textit{TopTrack} uses the \textit{Stacked Hourglass}\cite{StackedHourglass} backbone for feature extraction. Each hourglass module consists of 5 down and up-convolutional neural networks with skip connections \cite{ObjectsAsPoints}. Each network head consists of a $3\times3$ convolutional layer followed by a $1\times1$ convolutional layer to obtain the final prediction. All training was done using the Adam optimizer. The training was performed on a single RTX3090 GPU and an Intel Core i7-10700 CPU. Our model was first pretrained on the CrowdHuman\cite{CrowdHuman} dataset for 140 epochs with a batch size of 8 and a learning rate of $2.5^{-4}$. The starting learning rate is reduced by a factor of 10 after 90 epochs and once more after 120 epochs. We then further train the model on either the MOT17 dataset or MOT20 dataset to fine-tune it to the each respective benchmark. The MOT17 model was trained for an additional 70 epochs with a batch size of 32 and a learning rate of $1.25^{-4}$ reduced by a factor of 10 after 60 epochs. The MOT20 model was trained for an additional 90 epochs with a batch size of 32 and a learning rate of $2.5^{-4}$ reduced by a factor of 10 after 60 epochs. 

\subsection{Results}\label{sec2} 

\begin{table*}[h]
\begin{center}
\caption{Results of state-of-the-art methods on MOT17 dataset with private detections. Best results are in \textbf{bold}.}\label{tab1}%
\begin{tabular}{c | c c c c c c c c} 
\hline
Model & MOTA$\uparrow$ & HOTA$\uparrow$ & IDF1$\uparrow$ & MT$\uparrow$ & ML$\downarrow$ & FP$\downarrow$ & FN$\downarrow$ & IDSW$\downarrow$ \\ [0.4ex] 
\hline\hline
CenterTrack\cite{TrackingObjectsAsPoints} & 67.8 & 52.2 & 64.7 & 34.6 & 24.6 & \textbf{18 498} & 160 332 & 3039 \\ [0.4ex]
CTracker\cite{ChainedTracker} & 66.6 & 49.0 & 57.4 & 32.2 & 24.2 & 22 284 & 160 491 & 5529 \\
FairMOT\cite{FairMOT} & 73.7 & 59.3 & 72.3 & 43.2 & 17.3 & 27 507 & 117 477 & 3303 \\
GSDT\_v2\cite{GSDT} & 73.2 & 55.2 & 66.5 & 41.7 & 17.5 & 26 397 & 120 666 & 3891 \\
MOTR\cite{MOTR} & \textbf{78.6} & \textbf{62.0} & \textbf{75.0} & 50.3 & 13.1 & 23 409 & 94 797 & 2619 \\
SCSAN \cite{SCSAN} & 66.9 & - & 68.3 & 35.7 & 21.5 & 23 587 & 193 286 & \textbf{1868} \\
SST\cite{SST} & 52.4 & 39.3 & 49.5 & 18.3 & 34.0 & 25 423 & 234 592 & 8431 \\
TransCenter\cite{TransCenter} & 76.4 & - & 65.4 & \textbf{51.7} & 11.6 & 37 005 & \textbf{89 712} & 6402  \\
TubeTK\cite{TubeTK} & 63.0 & 48.0 & 58.6 & 31.2 & 19.9 & 27 060 & 177 483 & 4137 \\
\hline
Ours & 64.8 & 47.9 & 58.2 & 38.7 & \textbf{11.5} & 52 425 & 136 221 & 10 083 \\ 
\hline
\end{tabular}
\end{center}
\end{table*}

\begin{table*}[h]
\begin{center}
\caption{Results of state-of-the-art methods on MOT20 dataset with private detections. Best results are in \textbf{bold}.}\label{tab2}%
\begin{tabular}{c | c c c c c c c c} 
\hline
Model & MOTA$\uparrow$ & HOTA$\uparrow$ & IDF1$\uparrow$ & MT$\uparrow$ & ML$\downarrow$ & FP$\downarrow$ & FN$\downarrow$ & IDSW$\downarrow$ \\ [0.4ex] 
\hline\hline
FairMOT\cite{FairMOT} & 61.8 & \textbf{54.6} & 67.3 & \textbf{68.8} & \textbf{7.6} & 103 440 & \textbf{88 901} & 5243 \\
GSDT\_v2\cite{GSDT} & 67.1 & 53.6 & \textbf{67.5} & 53.1 & 13.2 & 31 507 & 135 935 & 9878 \\
MLT\cite{MLT} & 48.9 & 43.2 & 54.6 & 30.9 & 22.1 & 45 660 & 216 803 & \textbf{2187} \\
TMM\cite{TMM} & 43.3 & 36.2 & 45.2 & 17.6 & 26.3 & \textbf{27 953} & 262 406 & 2965 \\
TransCenter\cite{TransCenter} & \textbf{72.9} & - & 57.7 & 66.5 & 11.8 & 28 596 & 108 982 & 2625 \\
\hline
Ours & 46.3 & 26.8 & 27.6 & 20.0 & 22.0 & 28 363 & 226 089 & 23 227\\ 
\hline
\end{tabular}
\end{center}
\end{table*}

\begin{figure*}
\raisebox{0.85in}{\rotatebox[origin=c]{90}{\bfseries TopTrack\strut}}
    \includegraphics[width=0.5\textwidth]{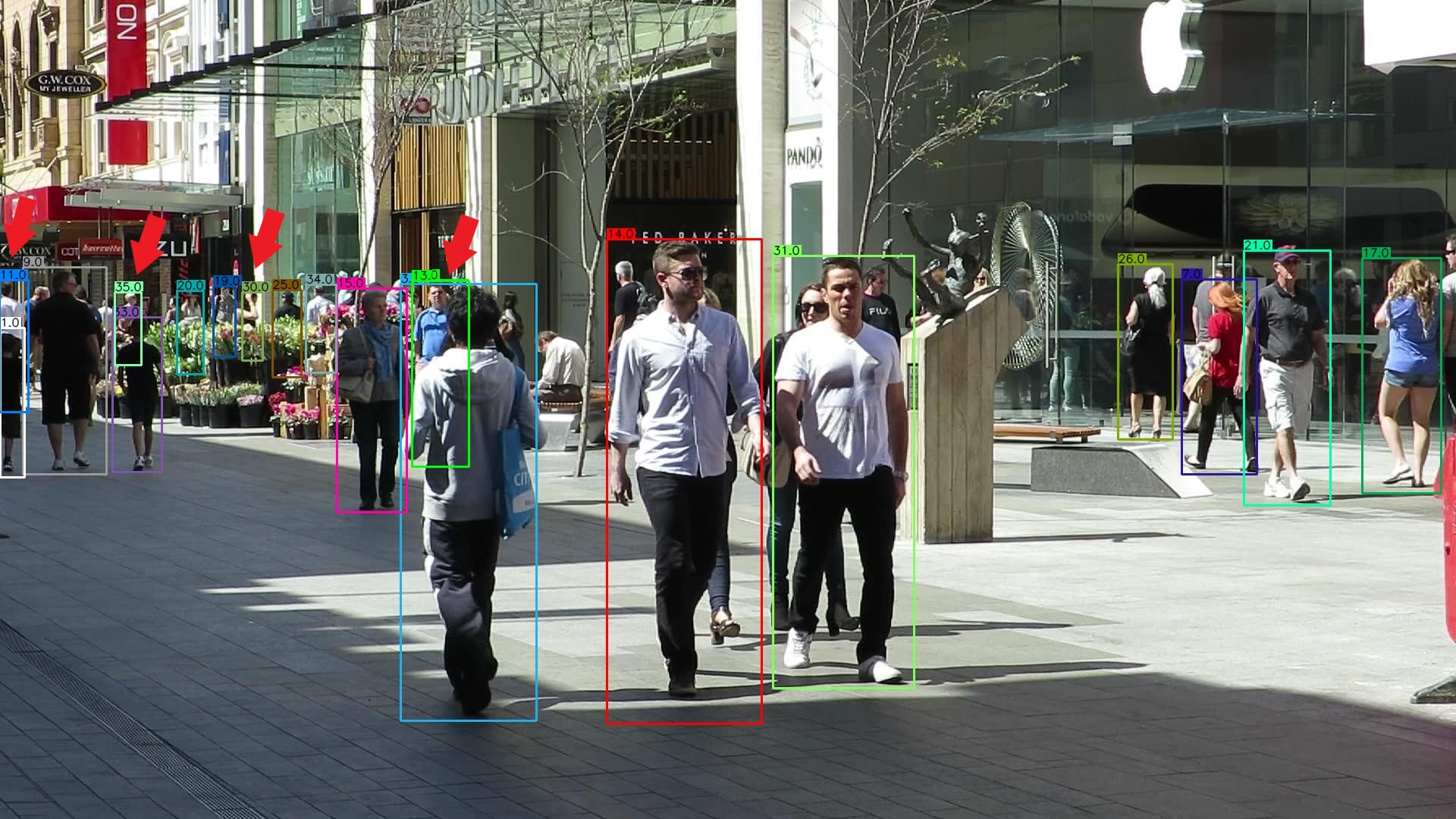}
    \includegraphics[width=0.5\textwidth]{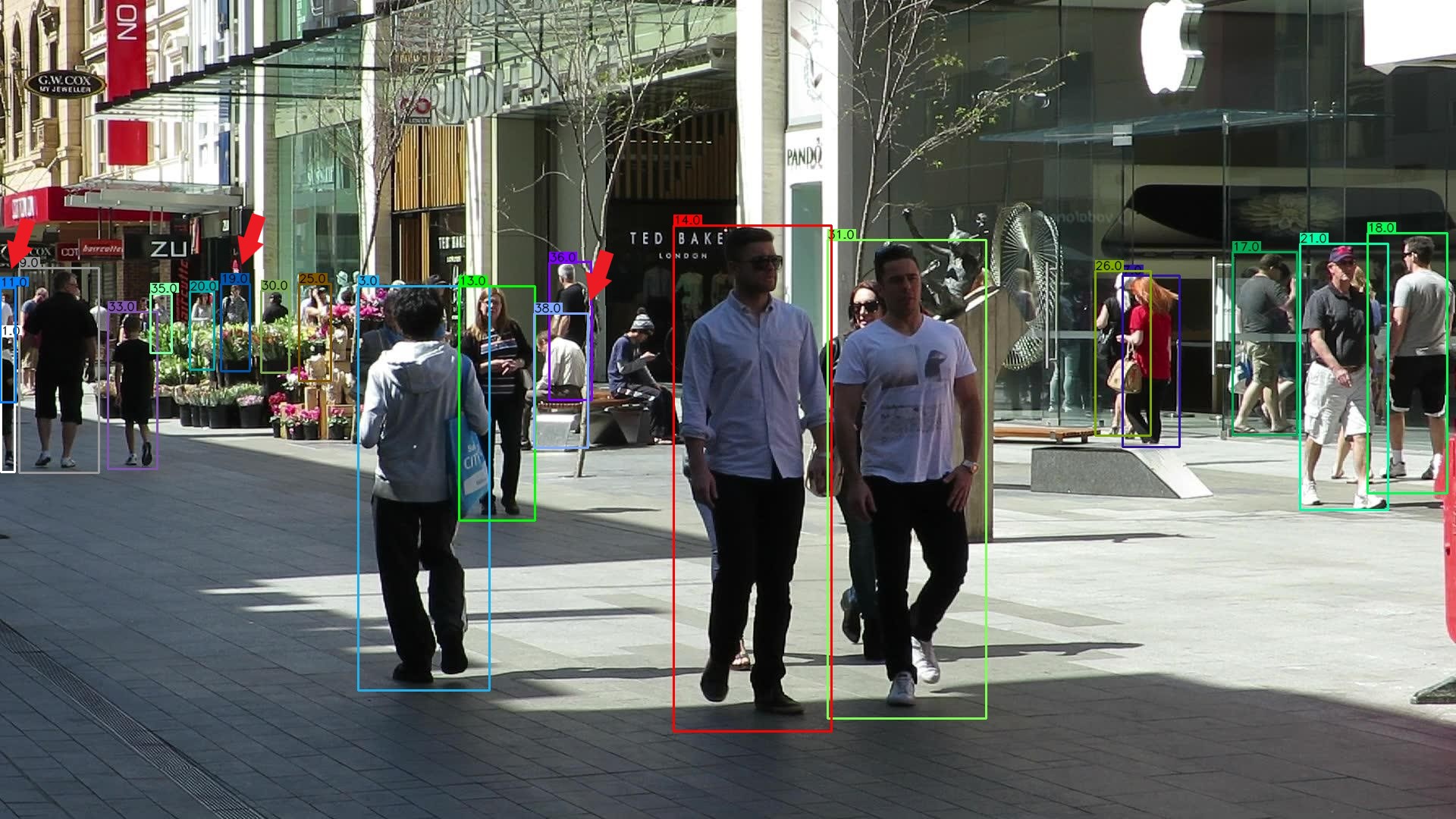}
\raisebox{0.85in}{\rotatebox[origin=c]{90}{\bfseries CenterTrack\strut}}
    \includegraphics[width=0.5\textwidth]{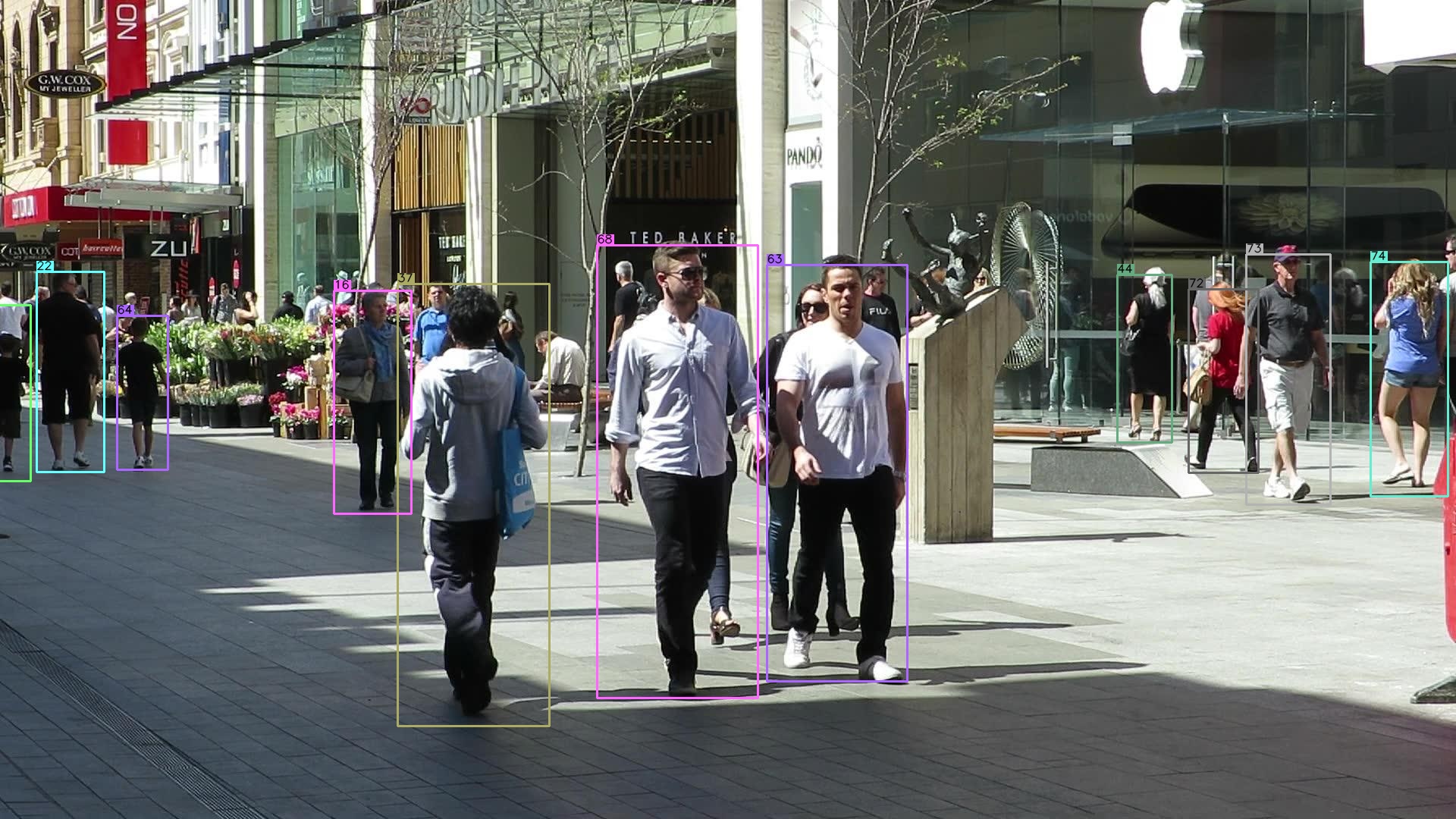}
    \includegraphics[width=0.5\textwidth]{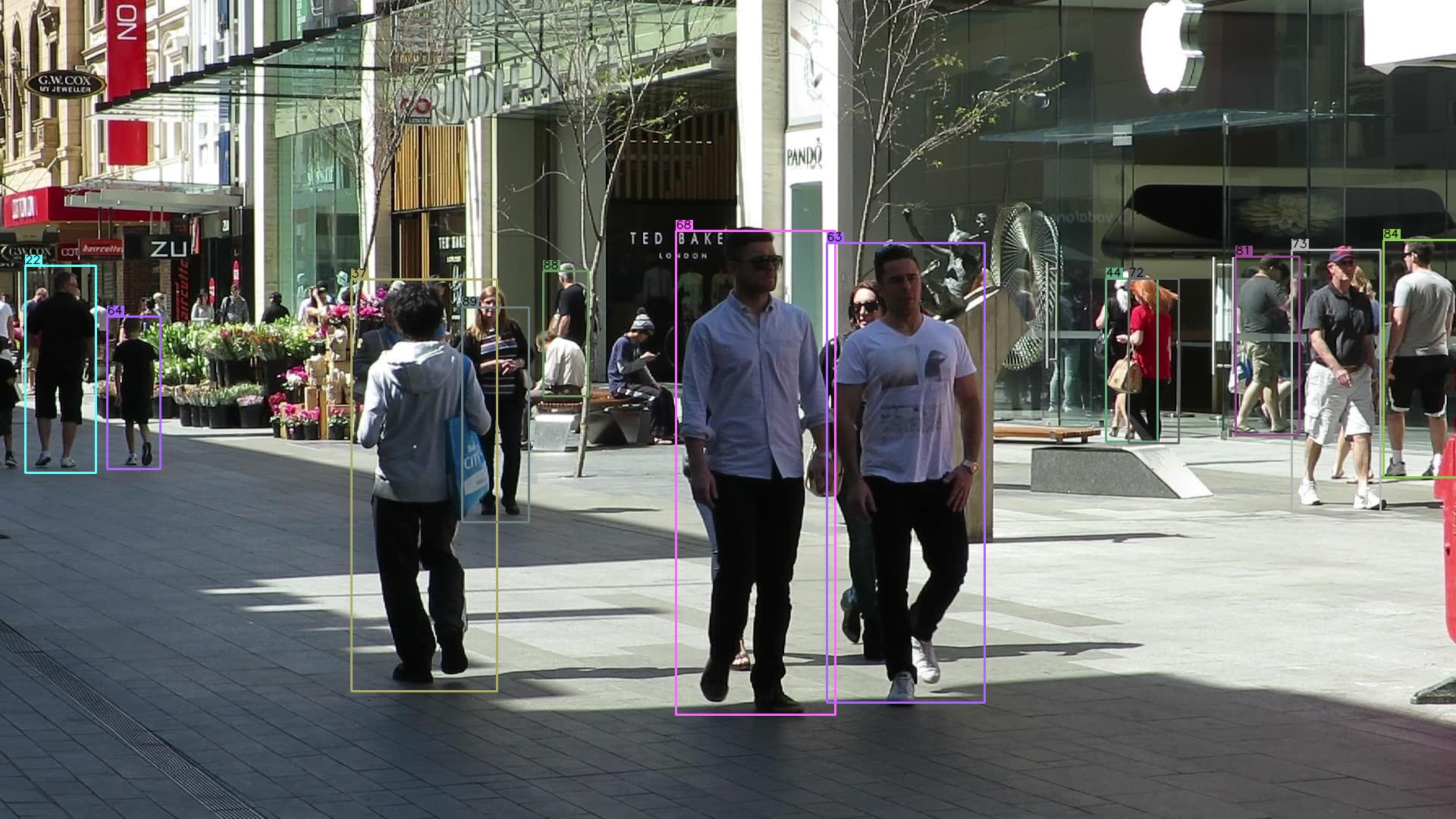}
\caption{Comparison between TopTrack and CenterTrack on two frames of sequence 8 of MOT17 test set. Arrows show objects detected by TopTrack but missed by CenterTrack}
\label{comparison1}
\end{figure*}

\begin{figure*}
\raisebox{0.85in}{\rotatebox[origin=c]{90}{\bfseries TopTrack\strut}}
    \includegraphics[width=0.5\textwidth]{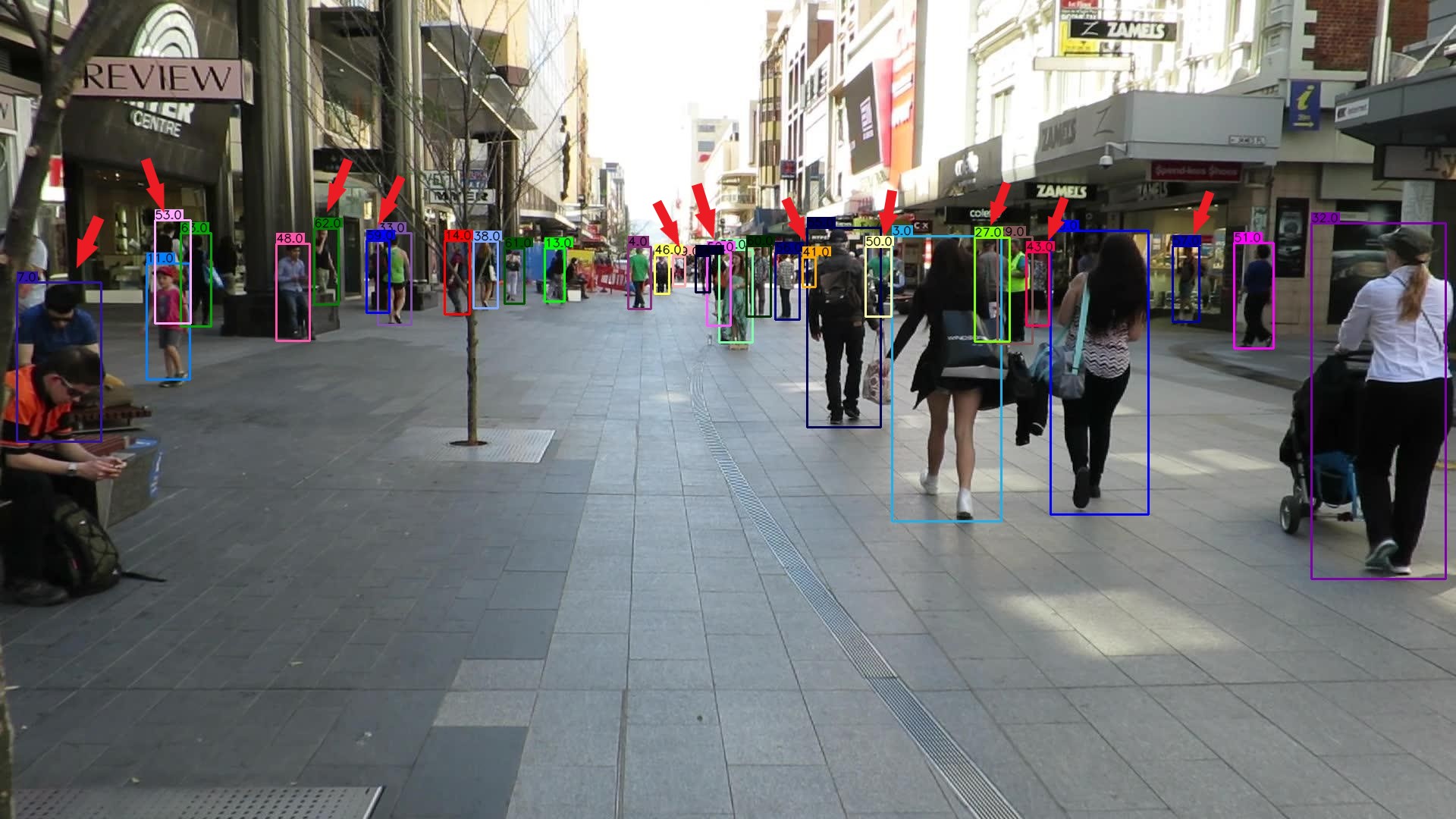}
    \includegraphics[width=0.5\textwidth]{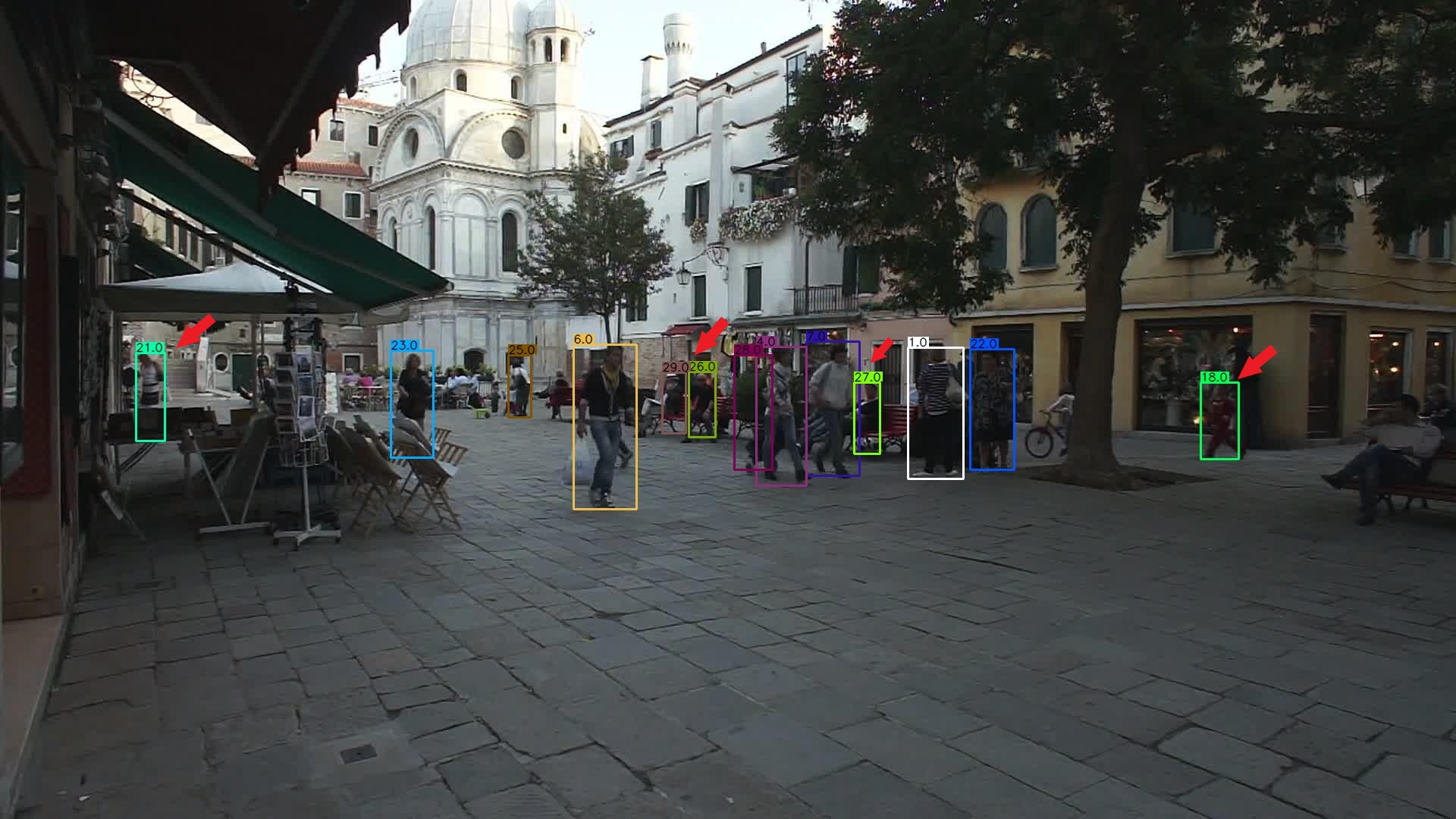}
\raisebox{0.85in}{\rotatebox[origin=c]{90}{\bfseries CenterTrack\strut}}
    \includegraphics[width=0.5\textwidth]{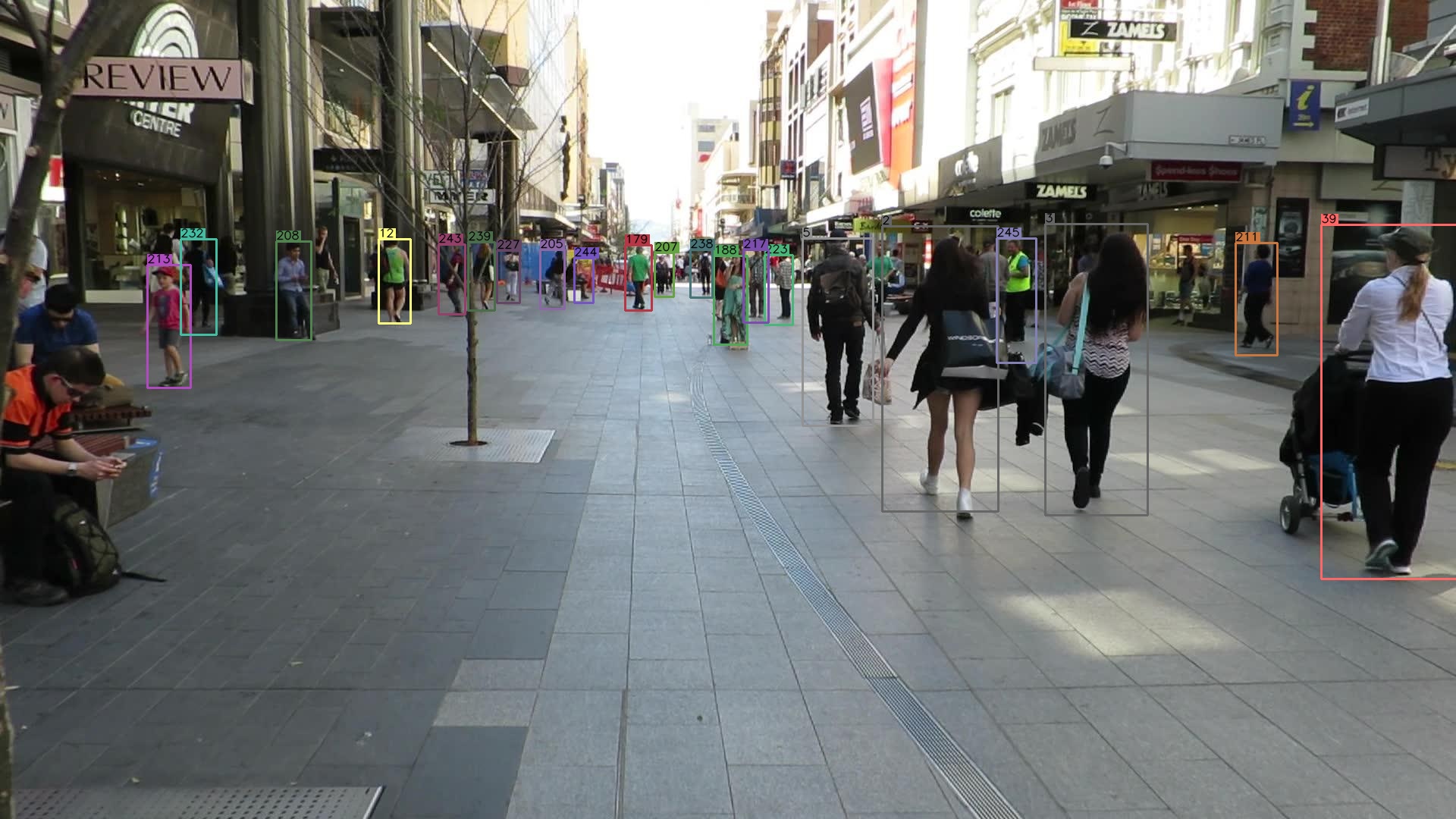}
    \includegraphics[width=0.5\textwidth]{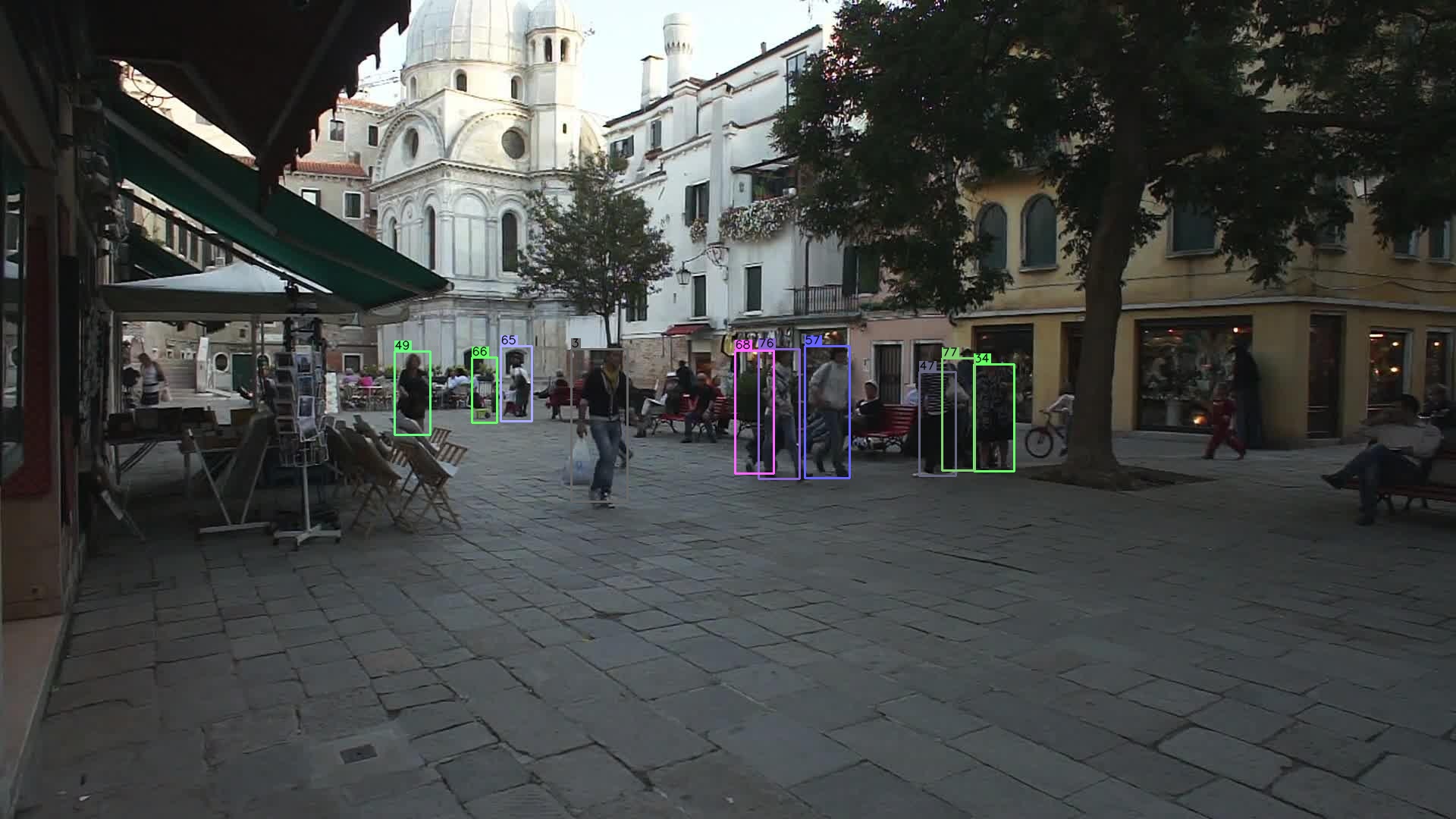}
\caption{Comparison between TopTrack and CenterTrack on one frame of sequence 7 and one frame of sequence 1 of MOT17 test set. Arrows show objects detected by TopTrack but missed by CenterTrack}
\label{comparison2}
\end{figure*}

As we can see from table \ref{tab1}, our overall results are slightly lower than \textit{CenterTrack} and other more recent trackers under the joint-tracking-and-detection paradigm. However, we did manage to significantly reduce the number of False Negative errors (missed detections), which supports the motivation of using object top information for better detection. Also, we managed to achieve the best score for the ML metric and a very high MT metric, meaning that our tracker manages to accurately track most trajectories for a significant amount of time and misses very few trajectories. Moreover, these trajectories that other trackers did not manage to track are often the most difficult examples, which goes to show that there is value in using the top of the objects as a keypoint for detection. As it can be seen in figure \ref{comparison1} and \ref{comparison2}, our trackers manages to detect and track many objects that other trackers cannot, like objects that are small and far away, that are partially occluded or that are sitting, which makes them a lot harder to detect.

\begin{figure*}
    \includegraphics[width=0.5\textwidth]{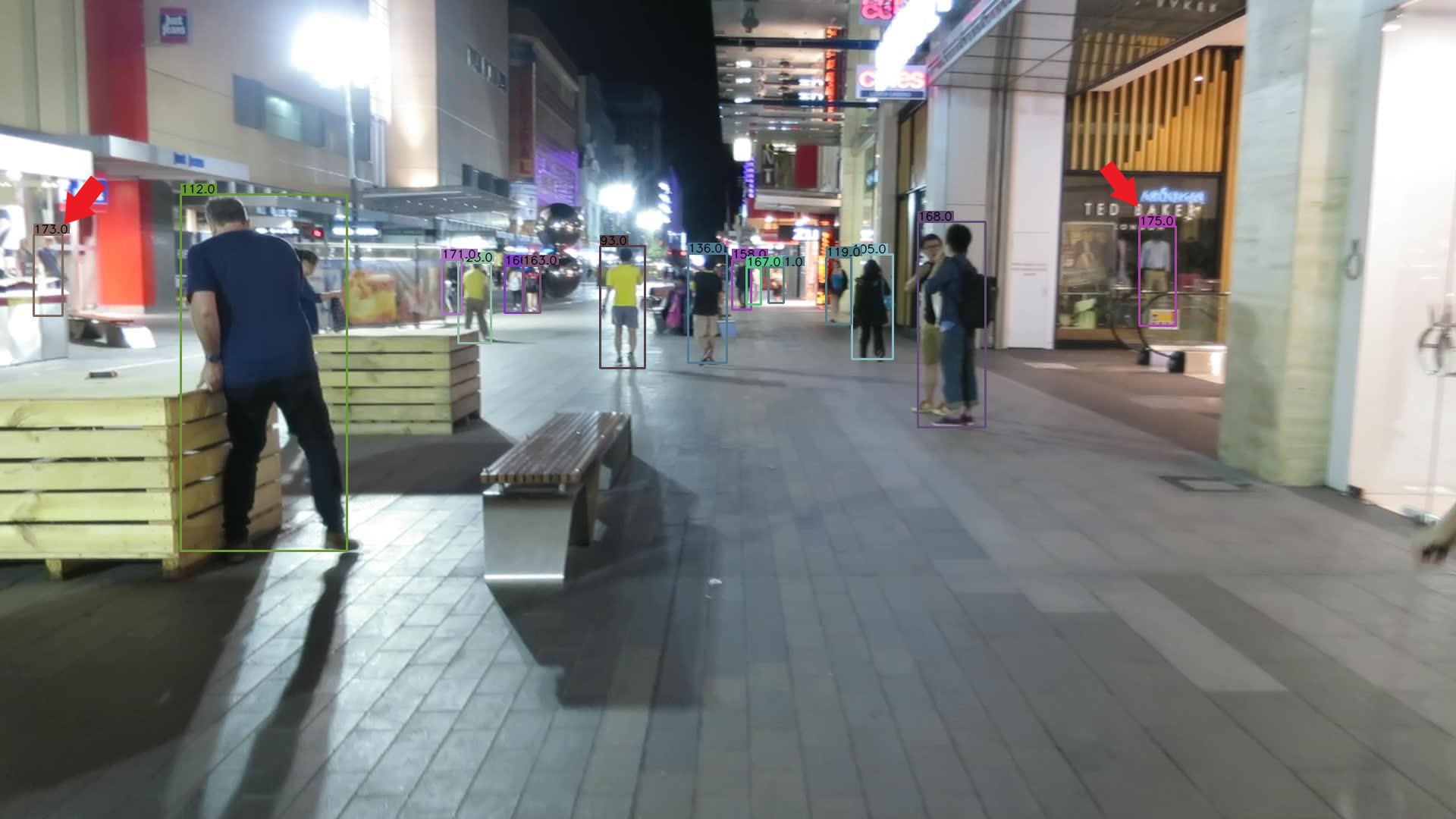} \includegraphics[width=0.5\textwidth]{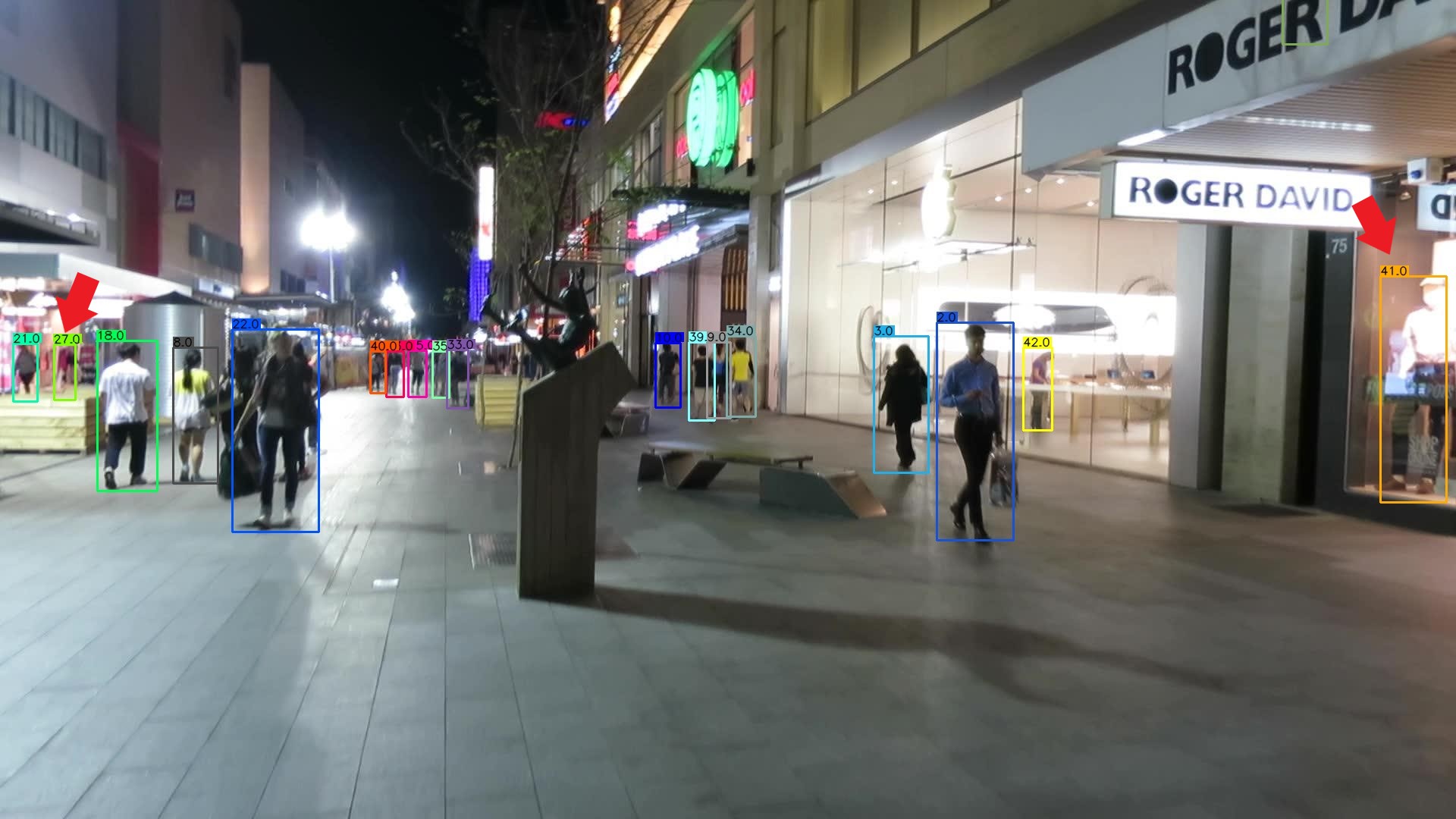}
    \caption{Examples of shop mannequins being detected as objects by TopTrack. See ID 173 and 175 (left) and ID 27 and 41 (right)}
    \label{fig:FP_mannequins}
\end{figure*}

\begin{figure*}
    \includegraphics[width=0.5\textwidth]{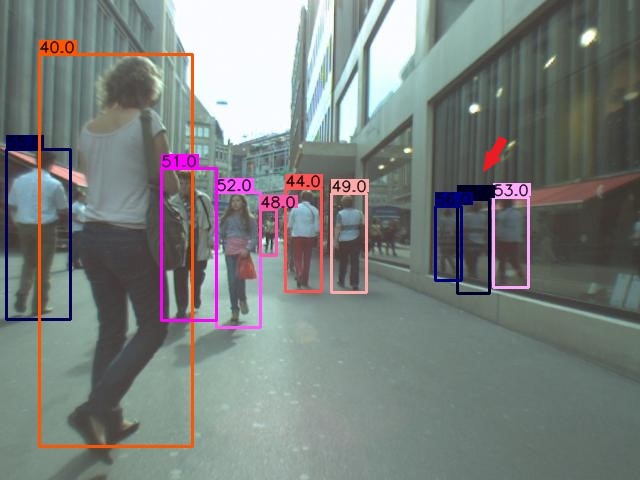} \includegraphics[width=0.5\textwidth]{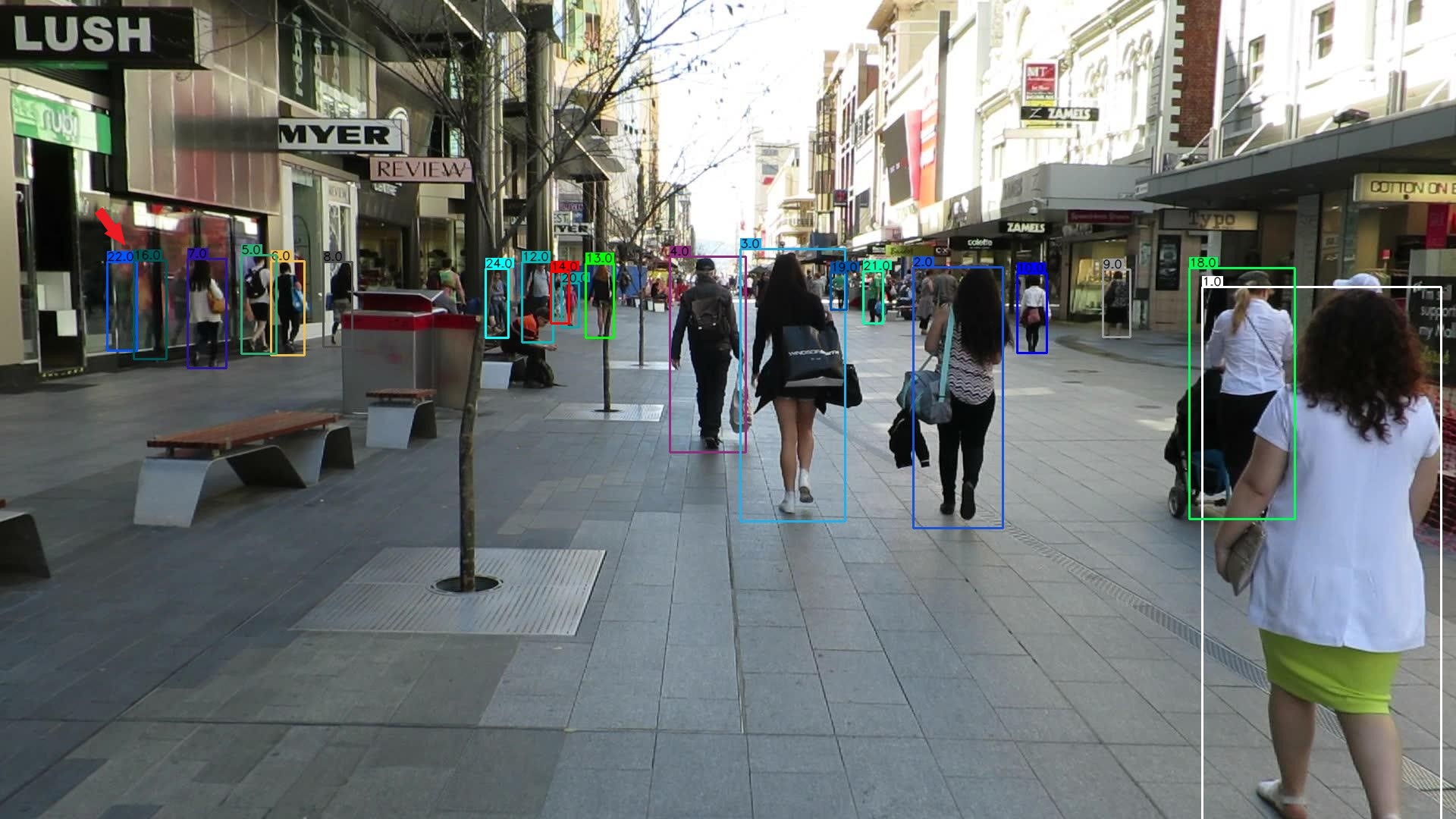}
    \caption{Examples of reflections being detected as objects by TopTrack. See ID 56, 63 and 64 (left) and ID 16 and 22 (right)}
    \label{fig:FP_reflections}
\end{figure*}

\begin{figure*}
\centering
    \includegraphics[width=0.5\textwidth]{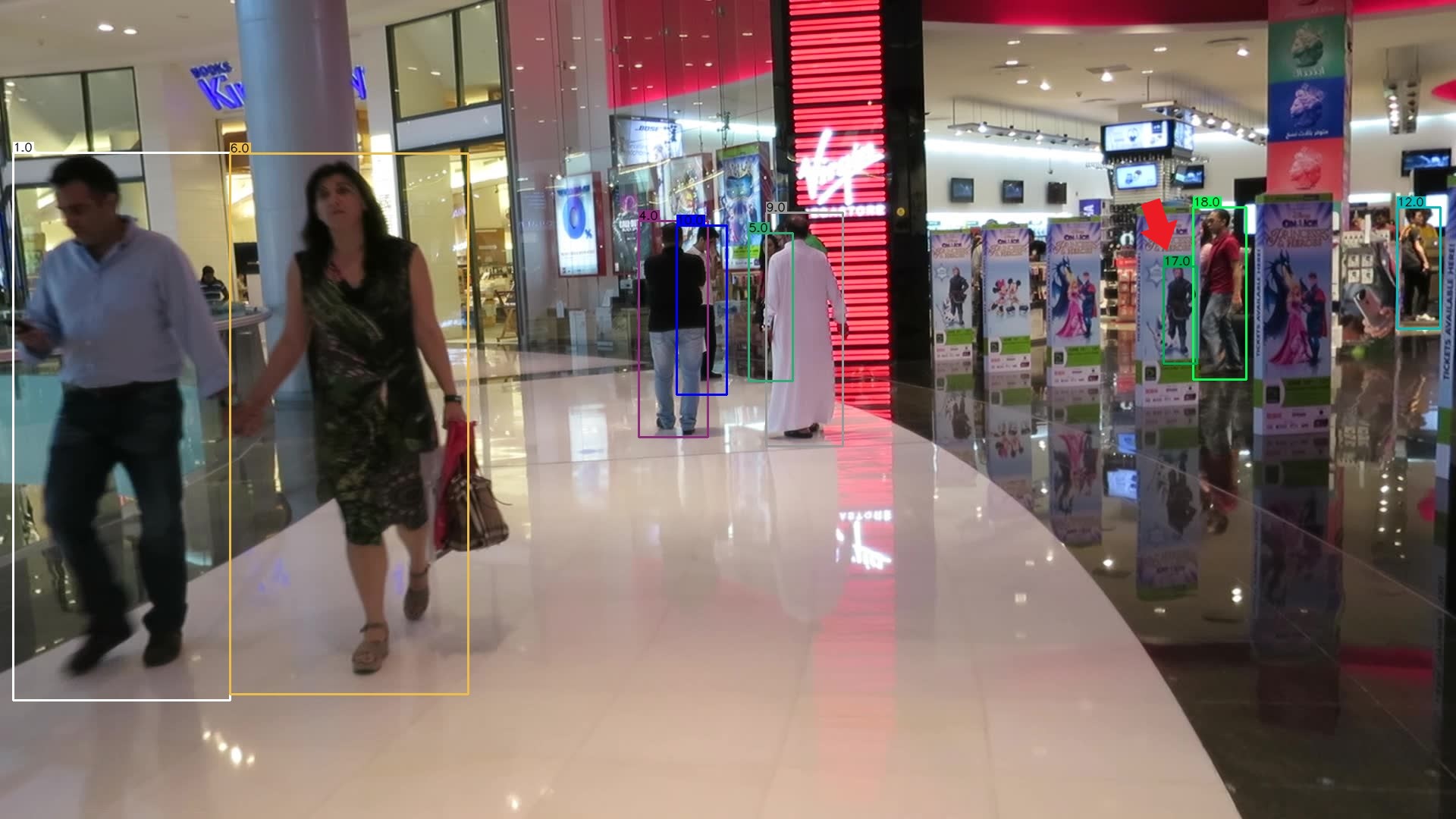}
    \caption{Examples of a cartoon character being detected as an object by TopTrack (See ID 17)}
    \label{fig:FP_cartoon}
\end{figure*}

Looking at the performance of \textit{TopTrack} on the test sequences, a high amount of FP can be attributed to human-like objects. Figures \ref{fig:FP_mannequins}, \ref{fig:FP_reflections} and \ref{fig:FP_cartoon} show how our tracker detects shop mannequins, reflections from the crowd in windows and cartoon characters on advertisements. Furthermore, objects that are not evaluated by the benchmarks like seated people and cyclists are also considered as FP errors even though it is reasonable for our model to detect them. We are confident that many of these errors could be alleviated with better training using more data as described in \cite{FairMOT} for example. The best models also all use many more datasets to train their model. As for the high amount of IDSW, this can mostly be attributed to the higher amount of detections from our tracker and the usage of a very simple data association strategy. Having more detections means there are more possibilities to make errors, especially for harder examples. Furthermore, like for \textit{CenterTrack}, our data association strategy is local only and does not use appearance information and does not allow for track rebirths. We think that adopting a more robust association scheme would greatly reduce the amount of IDSW, which would improve the overall score of our tracker. These factors are even more relevant when looking at table \ref{tab2}. Indeed, the sequences of the MOT20 challenge have very dense crowds meaning it is very hard for a local data association scheme to be accurate. 

\subsection{Ablation Study} \label{Ablation Study}

The most important aspect of our method to evaluate is the effect of using the top keypoint for detection instead of the center. To do so, we trained a model using each keypoint and present the results in Table \ref{hm_ablation}. Using the top keypoint reduces the amount of FP, FN and increases IDF1, which results in a better MOTA score.

\begin{table}[h]
\begin{center}
\caption{Results of TopTrack on the MOT17 validation set using keypoint for detection. Best results are in bold}\label{hm_ablation}
\vspace{-1\baselineskip}
\resizebox{\columnwidth}{!}{%
\begin{tabular}{@{\hspace{0.5\tabcolsep}}c@{\hspace{0.5\tabcolsep}}|@{\hspace{0.5\tabcolsep}}c@{\hspace{1\tabcolsep}}c@{\hspace{1\tabcolsep}}c@{\hspace{1\tabcolsep}}c@{\hspace{1\tabcolsep}}c@{\hspace{1\tabcolsep}}c@{\hspace{1\tabcolsep}}c@{\hspace{0.5\tabcolsep}}} 
\hline
Keypoint & MOTA$\uparrow$ & IDF1$\uparrow$ & MT$\uparrow$ & ML$\downarrow$ & FP$\downarrow$ & FN$\downarrow$ & IDSW$\downarrow$ \\ 
\hline\hline
Center & 58.0 & 60.2 & 29.8 & \textbf{21.5} & 5.9\% & 35.1\% & \textbf{0.9\%} \\
Top & \textbf{60.7} & \textbf{61.5} & \textbf{30.1} & 21.8 & \textbf{4.7\%} & \textbf{33.7\%} & \textbf{0.9\%} \\
\hline
\end{tabular}%
}
\end{center}
\end{table}

For data association, a common simple association scheme is the Hungarian algorithm. Table \ref{DA_ablation} shows how using it affects the results when compared to the greedy algorithm that we used. The main difference is a significant increase in IDSW when using the Hungarian algorithm and also a slight increase in FP and a slight decrease in FN. Overall, using the greedy algorithm gives better results.

\begin{table}[h]
\begin{center}
\caption{Results of TopTrack on the MOT17 validation set using different matching algorithms. Best results are in bold}\label{DA_ablation}
\vspace{-1\baselineskip}
\resizebox{\columnwidth}{!}{%
\begin{tabular}{@{\hspace{0.5\tabcolsep}}c@{\hspace{0.5\tabcolsep}}|@{\hspace{0.5\tabcolsep}}c@{\hspace{1\tabcolsep}}c@{\hspace{1\tabcolsep}}c@{\hspace{1\tabcolsep}}c@{\hspace{1\tabcolsep}}c@{\hspace{1\tabcolsep}}c@{\hspace{1\tabcolsep}}c@{\hspace{0.5\tabcolsep}}} 
\hline
Algorithm & MOTA$\uparrow$ & IDF1$\uparrow$ & MT$\uparrow$ & ML$\downarrow$ & FP$\downarrow$ & FN$\downarrow$ & IDSW$\downarrow$ \\ 
\hline\hline
Greedy & \textbf{60.7} & \textbf{61.5} & 30.1 & 21.8 & \textbf{4.7\%} & 33.7\% & \textbf{0.9\%} \\
Hungarian & 60.5 & 53.5 & \textbf{32.4} & \textbf{20.9} & 5.3\% & \textbf{33.0\%} & 1.2\% \\
\hline
\end{tabular}%
}
\end{center}
\end{table}

Finally, we tested our method using different training seeds. Seed 317 is the default used in \textit{CenterNet} \cite{ObjectsAsPoints} and \textit{CornerNet} \cite{CornerNet}. The other seeds were chosen randomly. As it can be seen in Table \ref{seed_ablation}, the default seed is the best one. Most other seeds however did not result in a significant decrease in accuracy except for seed 5. The model using this seed had a significant amount of FN more than the others. Globally, results are stable and not to sensitive to the choice of the seed. 

\begin{table}[h]
\begin{center}
\caption{Results of TopTrack on the MOT17 validation set using different seeds for training. Best results are in bold}\label{seed_ablation}
\vspace{-1\baselineskip}
\begin{tabular}{@{\hspace{0.5\tabcolsep}}c|c@{\hspace{1\tabcolsep}}c@{\hspace{1\tabcolsep}}c@{\hspace{1\tabcolsep}}c@{\hspace{1\tabcolsep}}c@{\hspace{1\tabcolsep}}c@{\hspace{1\tabcolsep}}c@{\hspace{0.5\tabcolsep}}} 
\hline
Seed & MOTA$\uparrow$ & IDF1$\uparrow$ & MT$\uparrow$ & ML$\downarrow$ & FP$\downarrow$ & FN$\downarrow$ & IDSW$\downarrow$ \\ 
\hline\hline
317 & \textbf{60.7} & 61.5 & 30.1 & \textbf{21.8} & 4.7\% & \textbf{33.7\%} & 0.9\% \\
310 & 60.2 & 60.7 & 28.0 & 27.7 & 4.5\% & 34.3\% & 0.9\% \\
17 & 60.3 & 61.3 & 28.7 & 27.4 & 4.7\% & 33.9\% & 0.9\% \\
5 & 58.0 & 59.4 & 27.1 & 26.8 & 4.8\% & 36.4\% & \textbf{0.8\%} \\
142 & 60.6 & \textbf{61.8} & \textbf{31.0} & 24.2 & \textbf{4.4\% }& 34.2\% & 0.9\% \\
\hline
\end{tabular}
\end{center}
\end{table}

\section{Conclusion}\label{sec13}

In this work, we introduced an end-to-end trainable joint-detection-and-tracking model that uses the top of the object as a keypoint for detection and tracking instead of the more commonly used center point. We show how using this keypoint allows our tracker to better detect hard-to-track objects and we also show how using the generated heatmap of the current frame allows the tracker to better connect objects locally, which leads to a reduction in FN errors. Our model achieves good accuracy and competitive results on two popular MOT benchmarks.

\backmatter

\section*{Statements and Declarations}

\bmhead{Funding}
This work was supported by the Natural Sciences and Engineering Research Council of Canada (NSERC), [funding reference number DGDND-2020-04633].

\bmhead{Conflict of interest}
The authors have no conflict of interests to disclose.

\bmhead{Ethics approval}
Not applicable.

\bmhead{Consent to participate}
Not applicable.

\bmhead{Consent for publication}
All authors consent to the publication of this article.

\bmhead{Availability of data and materials}
All the datasets that were used for this work are freely available online on their respective official website. MOT17 is available at \url{https://motchallenge.net/data/MOT17/}. MOT20 is available at \url{https://motchallenge.net/data/MOT20/}. CrowdHuman is available at \url{https://www.crowdhuman.org/}.

\bmhead{Code availability}
The code used for this work is freely available online at the following URL: \url{https://github.com/TopTrack/TopTrack2023}

\bmhead{Authors' contributions}
Jacob Meilleur conducted the research project and wrote the first draft of the manuscript. Guillaume-Alexandre Bilodeau supervised the research project, revised the manuscript and secured the funding needed to conduct this work.

\bibliographystyle{IEEEtran}
\bibliography{sn-bibliography}


\end{document}